\documentclass[runningheads]{llncs}
  \usepackage{graphicx}
  \usepackage{amsmath,amssymb} 
  \usepackage{color}
  
  \usepackage{bm}
  \usepackage{subfigure}
  \usepackage{booktabs}
  \usepackage{tabularx}
  \usepackage{multirow}
  \usepackage{array}
  \usepackage{verbatim}
  \usepackage{enumerate}
  \usepackage{anyfontsize}
  \usepackage{textgreek}
  \usepackage{overpic}
  \usepackage[colorlinks, linkcolor=blue, anchorcolor=blue,  citecolor=blue]{hyperref}
  \usepackage{enumitem}
  
  \newcommand{\PreserveBackslash}[1]{\let\temp=\\#1\let\\=\temp}
  \newcommand\norm[1]{\left\lVert#1\right\rVert}

  \newcolumntype{C}[1]{>{\PreserveBackslash\centering}p{#1}}
  \newcolumntype{R}[1]{>{\PreserveBackslash\raggedleft}p{#1}}
  \newcolumntype{L}[1]{>{\PreserveBackslash\raggedright}p{#1}}

  \def\red{\color{red}}
  
\begin{document}
  \title{Online Multi-Object Tracking with\\Dual Matching Attention Networks} 
  \titlerunning{Online MOT with Dual Matching Attention Networks}
  %
  \author{Ji Zhu$^{1,2}$, Hua Yang$^{1\star}$, Nian Liu$^3$, Minyoung Kim$^4$,\\Wenjun Zhang$^1$, and Ming-Hsuan Yang${^{5,6}}$}
  \authorrunning{J. Zhu, H. Yang, N. Liu, M. Kim, W. Zhang, and M.-H. Yang}
  \institute{$^1$Shanghai Jiao Tong University~~~$^2$Visbody Inc\\
  $^3$Northwestern Polytechnical University~~~$^4$Massachusetts Institute of Technology\\
  $^5$University of California, Merced~~~$^6$Google Inc\\
  \email{\{jizhu1023, liunian228\}@gmail.com}~~~\email{minykim@mit.edu}\\
  \email{\{hyang,zhangwenjun\}@sjtu.edu.cn}~~~\email{mhyang@ucmerced.edu}}
  \footnotetext[1]{Corresponding author.}
  \maketitle              
  \begin{abstract}
  In this paper, we propose an online Multi-Object Tracking (MOT) approach which integrates the merits of single object tracking and data association methods in a unified framework to handle noisy detections and frequent interactions between targets. 
  Specifically, for applying single object tracking in MOT, we introduce a cost-sensitive tracking loss based on the state-of-the-art visual tracker, which encourages the model to focus on hard negative distractors during online learning. 
  For data association, we propose Dual Matching Attention Networks (DMAN) with both spatial and temporal attention mechanisms. 
  The spatial attention module generates dual attention maps which enable the network to focus on the matching patterns of the input image pair, while the temporal attention module adaptively allocates different levels of attention to different samples in the tracklet to suppress noisy observations.
  Experimental results on the MOT benchmark datasets show that the proposed algorithm performs favorably against both online and offline trackers in terms of identity-preserving metrics.
  
  \end{abstract}
  \section{Introduction}
  Multi-Object Tracking (MOT) aims to estimate trajectories of multiple objects by finding target locations and maintaining target identities across frames. 
  %
  %
  In general, existing MOT methods can be categorized into offline and online methods. 
  Offline MOT methods use both past and future frames to generate trajectories while online MOT methods only exploit the information available up to the current frame. 
  Although offline methods have some advantages in handling ambiguous tracking results, they are not applicable to real-time vision tasks.
  
  Recent MOT methods mainly adopt the tracking-by-detection strategy and handle the task by linking detections across frames using data association algorithms. 
  However, these approaches heavily rely on the quality of detection results. 
  If the detection is missing or inaccurate, the target object is prone to be lost. 
  To alleviate such issues, recent methods \cite{MDP,STAM} exploit single object tracking methods for MOT. 
  A single object tracker uses the detection in the first frame and online updates the model to find the target in following frames.
  However, it is prone to drift when the target is occluded.
  In this paper, we combine the merits of single object tracking and data association in a unified framework. 
  In most frames, a single object tracker is used to track each target object.
  Data association is applied when the tracking score is below a threshold, which indicates the target object may be occluded or undergo large appearance changes.

  The main challenge to use a single object tracker for MOT is to cope with frequent interactions between targets and intra-class distractors.
  Existing single object tracking methods usually suffer from the data imbalance issue between positive and negative samples for online model updates. 
  In the search area of a tracker, only a few locations near the target center correspond to positive samples while all the samples drawn at other positions are negative samples.
  Most locations from the background region are easy negatives, which may cause inefficient training and weaken the discriminative strength of the model. 
  This problem is exacerbated in the context of MOT task. 
  If a model is overwhelmed by the easy background negatives, the tracker is prone to drift when similar distractors appear in the search area. 
  Thus, it is imperative to focus on a small number of hard examples during online updates to alleviate the drifting problems. 
  
\begin{figure}[t]
    \centering
    \includegraphics[width=0.9\textwidth]{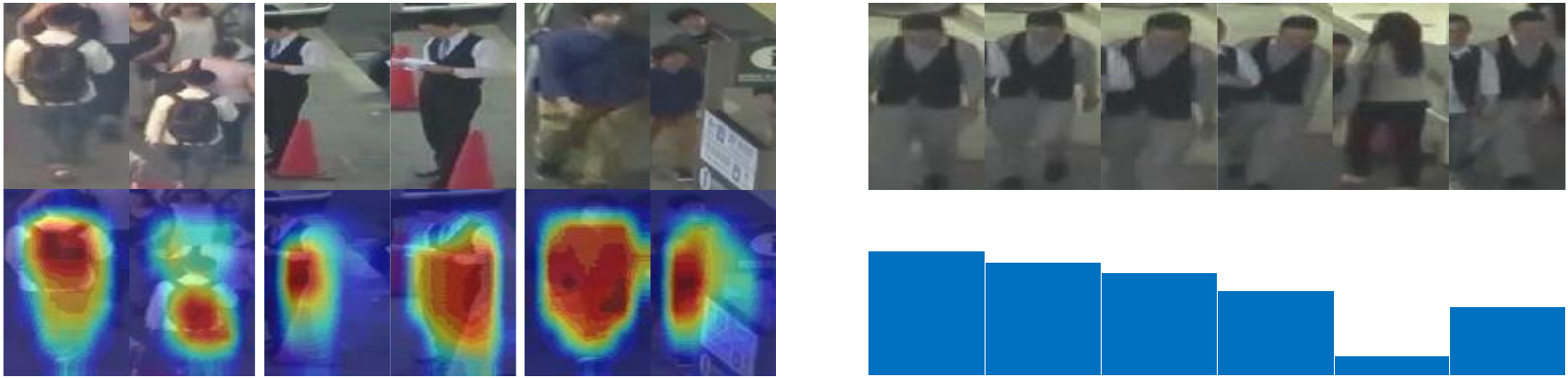}\\
    \begin{tabular}{@{}C{0.51\textwidth}C{0.5\textwidth}@{}}
    (a) & (b)\\
    \end{tabular}
    \caption{Sample detections in the MOT16 dataset \cite{MOT16}. (a) Top row: Image pairs with misalignments, missing parts, and occlusion. 
    Bottom row: Spatial attention maps for each image pair. (b) Top row: Target trajectory containing noisy samples. Bottom row: Temporal attention weights for corresponding images in the trajectory.}
    \label{fig:motivation}
  \end{figure}

  For data association, we need to compare the current detected target with a sequence of previous observations in the trajectory. 
  One of the most commonly tracked objects in MOT is pedestrian where the data association problem is also known as re-identification with challenging factors including pose variation, similar appearance, and frequent occlusion.
  In numerous public person re-identification datasets (e.g., \cite{CUHK01,CUHK02,CUHK03}), pedestrians given by manually annotated bounding boxes are well separated. 
  However, detected regions in the context of MOT may be noisy with large misalignment errors or missing parts 
  as shown in Fig.~\ref{fig:motivation}(a). 
  Furthermore, inaccurate and occluded observations in the previous trajectory likely result in noisy updates and make the appearance model less effective. 
  These factors motivate us to design an appearance model for effective data association in two aspects.
  First, to cope with misaligned and missing parts in detections, the proposed model should focus on corresponding local regions between observations, as presented in Fig.~\ref{fig:motivation}(a).
  Second, to avoid being affected by contaminated samples, the proposed model should assign different weights to different observations in the trajectory, as shown in Fig.~\ref{fig:motivation}(b).
  
  We make the following contributions in this work:
  \vspace{-3mm}
  \begin{itemize}[label=$\bullet$]
  \item We propose a spatial attention network to handle noisy detections and occlusions for MOT. 
  When comparing two images, the proposed network generates dual spatial attention maps (as shown in Fig.~\ref{fig:motivation}(a)) based on the cross similarity between each location of the image pair, which enables the model to focus on matching regions between the paired images without any part-level correspondence annotation.
  \item We design a temporal attention network to adaptively allocate different degrees of attention to different observations in the trajectory. 
  This module considers not only the similarity between the target detection and the observations in the trajectory but also the consistency of all observations to filter out unreliable samples in the trajectory.
  \item We apply the single object tracker in MOT and introduce a novel cost-sensitive tracking loss based on the state-of-the-art tracker. 
  The proposed loss enables the tracker to focus training on a sparse set of hard samples which enhances the robustness to nearby distractors in MOT scenarios.
  \item We carry out extensive experiments against the state-of-the-art MOT methods on the MOT benchmark datasets with 
  ablation studies to demonstrate the effectiveness of the proposed algorithm.
  \end{itemize}
  
  \section{Related Work}
  \subsubsection{Multi-Object Tracking.}
  Existing MOT methods tackle the task by linking the detections across consecutive frames based on the tracking-by-detection paradigm. 
  Numerous approaches \cite{milan2014continuous,pirsiavash2011globally,zhang2008global,JMC,QuadMOT,wang2016tracking,LMP} use detections from past and future frames for batch processing. 
  Typically, these methods model the MOT task as a global optimization problem in various forms such as 
  network flow \cite{zhang2008global,wang2016tracking,dehghan2015target}, and multi-cut \cite{JMC,LMP,tang2015subgraph}. 
  In contrast, online MOT methods \cite{MDP,STAM,leal2016learning} do not rely on detections from future frames and may not perform well when target objects are heavily occluded or mis-detected.  
  Thus, a robust appearance model is crucial for associating detections for online MOT. 
  Recently, several online approaches \cite{STAM,leal2016learning,milan2017online,AMIR,CDA_DDAL} using deep learning models have been proposed. 
  Leal-Taix{\'e} et al. \cite{leal2016learning} adopt a Siamese CNN to learn local features from both RGB images and optical flow maps. 
  In \cite{AMIR}, Sadeghian et al. propose to exploit the LSTM network to account for appearance modeling, which takes images in the tracklet step-by-step and predicts the similarity score. 
  In this work, we introduce attention mechanisms to handle inaccurate detections and occlusions. 
  We show that the proposed online algorithm achieves favorable identity-preserving performance against the state-of-the-art offline methods, even though the offline methods have the advantage of exploiting global information across frames. 
  
  \subsubsection{Attention Model.}
  A number of methods adopt attention mechanisms for various tasks such as image captioning \cite{chen2015mind,fang2015captions,xu2015show}, visual question answering \cite{xu2016ask,yang2016stacked}, and image classification \cite{wang2017residual}. 
  A visual attention mechanism enables the model to focus on the most relevant regions of the input to extract more discriminative features. 
  In this work, we integrate both spatial and temporal attention mechanisms into the proposed MOT algorithm. 
  Our approach differs from the state-of-the-art STAM metohd \cite{STAM}, which adopts the spatial-temporal attention mechanism for online MOT, in three aspects.
  First, the spatial attention in the STAM corresponds to the visibility map. Since the visibility map is estimated directly from the  detected image patch without comparison with the observations in the tracklet, it becomes unreliable when a distractor is close to the target. 
  In contrast, we exploit the interplay of the detection and tracklet to generate dual spatial attention maps, which is demonstrated to be more robust to noisy detections and occlusions.
  Second, the STAM needs to synthetically generate occluded samples and the corresponding ground truth to initialize model training while our spatial attention map can be learned implicitly without any pixel-level annotation. 
  Third, as the temporal attention value in \cite{STAM} is generated independently for each sample in the tracklet based on the estimated occlusion status, it is less effective when the distractor appears in the tracklet. We take the consistency of the overall tracklet into account and assign a lower attention weight to a noisy sample that is different from most samples in the tracklet.
  \subsubsection{Data Imbalance.}
  Data imbalance exists in numerous computer vision tasks where one class contains much fewer samples than others, which causes issues in training classifiers or model updates. 
  One common solution \cite{felzenszwalb2010cascade,shrivastava2016training} is to adopt hard negative mining during training. 
  Recently, several methods \cite{bulo2017loss,focal_loss} re-weight the contribution of each sample based on the observed loss 
  and demonstrate significant improvements on segmentation and detection tasks. 
  In this work, we propose a cost-sensitive tracking loss which puts 
  more emphasis on hard samples with large loss to alleviate drifting problems.
  \begin{figure}[t]
    \centering
    \includegraphics[width=0.9\textwidth]{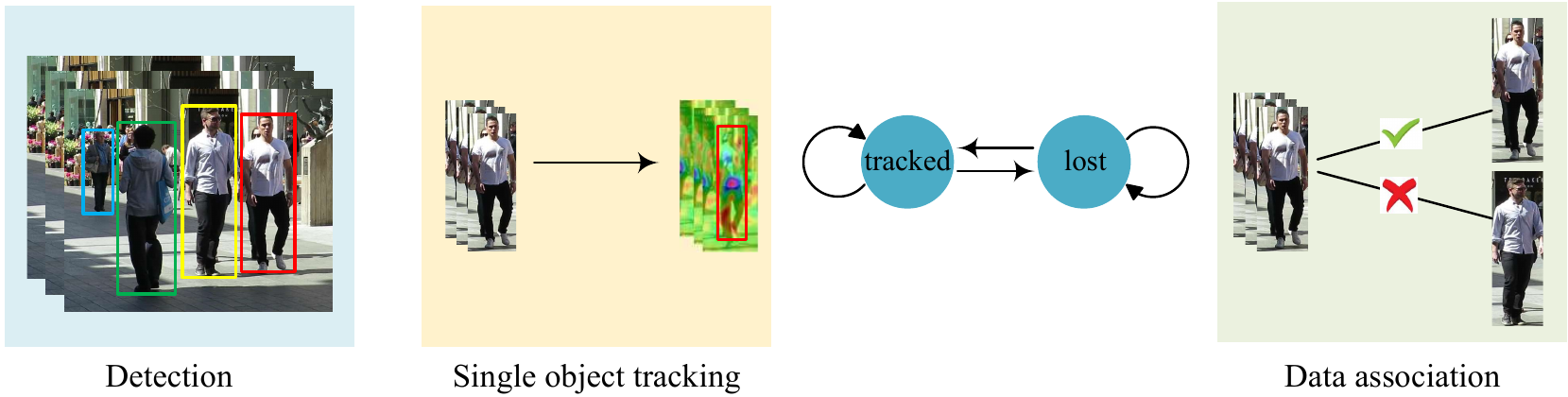}\\
    \caption{Proposed online MOT pipeline. This pipeline mainly consists of three tasks: detection, single object tracking, and data association. The state of each target switches between tracked and lost depending on the tracking reliability. Single object tracking is applied to generate the tracklets for the tracked targets while data association compares the tracklets with candidate detections to make assignments for the lost targets.}
  \label{fig:framework}
  \end{figure}
  
  \section{Proposed Online MOT Algorithm}
  We exploit both single object tracking and data association to maintain target identities. Fig.~\ref{fig:framework} illustrates the proposed online MOT pipeline.
  Given target detections in each frame, we apply a single object tracker to keep tracking each target. 
  The target state is set as tracked until the tracking result becomes unreliable (e.g., the tracking score is low or the tracking result is inconsistent with the detection result).
  In such a case, the target is regarded as lost. 
  We then suspend the tracker and perform data association to compute the similarity between the tracklet and detections that are not covered by any tracked target. 
  Once the lost target is linked to a detection through data association, we update the state as tracked and restore the tracking process.
  
  \vspace{-2mm}
  \subsection{Single Object Tracking}
  Since significant progress has been made on single object tracking in recent years, we apply the state-of-the-art single object tracker in MOT. 
  However, the tracker is prone to drift due to frequent interactions between different objects. 
  To alleviate this problem,  we propose a cost-sensitive tracking loss. 
  
  \vspace{-2mm}
  \subsubsection{Baseline Tracker.}
  We employ the method based on the Efficient Convolution Operators (ECO) \cite{ECO} as the baseline tracker. 
  The ECO tracker achieves the state-of-the-art performance on visual tracking benchmarks \cite{VOT,OTB,UAV,TC} and its fast variant ECO-HC based on hand-crafted features (HOG \cite{HOG} and Color Names \cite{CN}) operates at 60 frames per second (FPS) on a single CPU, which is suitable for the online MOT task.
  
  We first briefly review the ECO formulation as it is used as part of the proposed MOT algorithm.
  For clarity, we present the one-dimension domain formulation like \cite{ECO,CCOT}. Denote $\mathbf{x}=\{(\mathbf{x}^1)^\top, \cdots, (\mathbf{x}^D)^\top\}$ as a feature map with $D$ feature channels extracted from an image patch. Each feature channel $\mathbf{x}^d\in \mathbb{R}^{N_d}$ has a resolution $N_d$.
  Different from conventional correlation filter based trackers, the ECO tracker interpolates the discrete feature channel $\mathbf{x}^d$ to the continuous domain $[0,T)$ and aims to learn a continuous $T$-periodic multi-channel convolution filter $f=\{f^1,\cdots,f^D\}$ from a batch of $M$ training samples $\{\mathbf{x}_j\}_1^M$ by minimizing the following objective function:
  \begin{equation}\label{eq:obj_spatial}
  E(f)=\sum_{j=1}^M\alpha_j\norm{S_f\{\mathbf{x}_j\}(t) - y_j(t)}_{L^2}+\sum_{d=1}^D\norm{w(t)f^d(t)}_{L^2}, \quad t\in[0,T).
  \end{equation}
  Here, the factor $\alpha_j$ denotes the weight of the sample $\mathbf{x}_j$. The convolution operator $S_f$ maps the sample $\mathbf{x}_j$ to a score function $S_f\{\mathbf{x}_j\}(t)$, which predicts the confidence score of the target at the location $t\in[0,T)$ in the image. The label function $y_j(t)$ is the desired output of the operator $S_f$ applied to $\mathbf{x}_j$. The regularization term uses a weight function $w(t)$ to suppress boundary effects.
  
  The objective function \eqref{eq:obj_spatial} can be transformed to a least squares problem in the Fourier domain, which is equivalent to solve the following normal equation:
  \begin{equation}\label{eq:normal_eq_eco}
  (\mathbf{A}^{\mathrm{H}}\mathbf{\Gamma} \mathbf{A}+\mathbf{W}^\mathrm{H}\mathbf{W})\,\hat{\mathbf{f}}=\mathbf{A}^\mathrm{H}\mathbf{\Gamma} \hat{\mathbf{y}}.
  \end{equation}
  Here, the superscript $^\mathrm{H}$ denotes the conjugate-transpose of a matrix. We let $\hat{\mathbf{f}}=[(\hat{\mathbf{f}}^1)^\top, \cdots, (\hat{\mathbf{f}}^D)^\top]^\top$ denote the non-zero Fourier coefficient vector of the filter $f$, and let $\hat{\mathbf{y}}$ denote the corresponding label vector in the Fourier domain. The diagonal matrix $\mathbf{\Gamma}=\alpha_1\mathbf{I}\oplus\cdots\alpha_M\mathbf{I}$ contains the weight $\alpha_j$ for each sample $\mathbf{x}_j$. The matrix $\mathbf{A}=[(\mathbf{A}_1)^\top,\cdots, (\mathbf{A}_M)^\top]^\top$ is computed from the values of samples $\{\mathbf{x}_j\}_1^M$, while the block-diagonal matrix $\mathbf{W}=\mathbf{W}^1\oplus\cdots\mathbf{W}^D$ corresponds to the penalty function $w$ in \eqref{eq:obj_spatial}. 
  More details can be found in \cite{ECO,CCOT}.
  \begin{figure}[t]\label{fig:class_imbalance}
  \begin{minipage}[t]{0.5\textwidth}
  \centering
  \subfigure[]{
  \label{fig:class_imbalance_a}
  \includegraphics[width=0.7\textwidth]{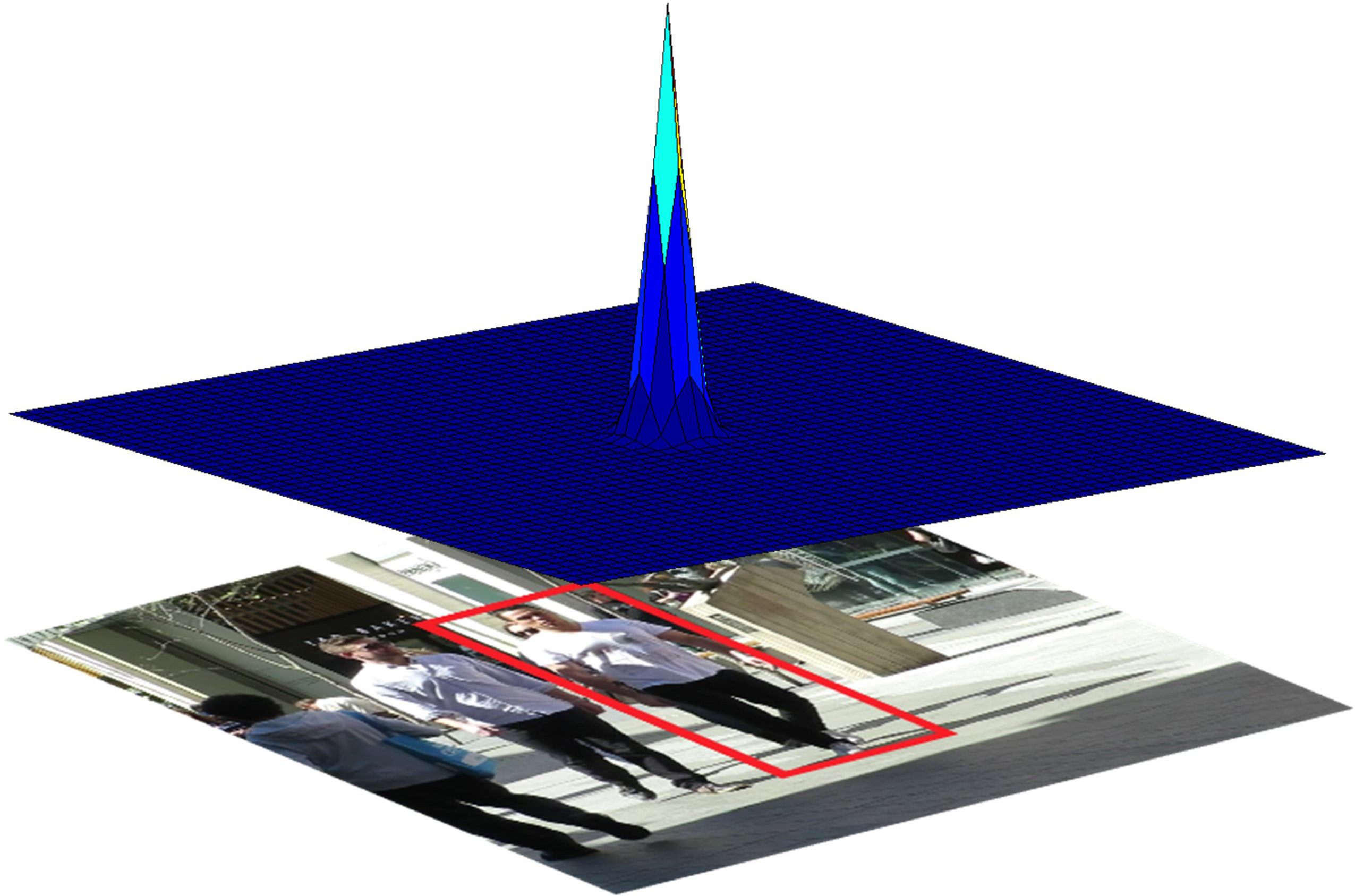}}\\
  \end{minipage}%
  \begin{minipage}[t]{0.5\textwidth}
  \centering
   \subfigure[]{
    \label{fig:class_imbalance_b}
    \includegraphics[width=0.7\textwidth]{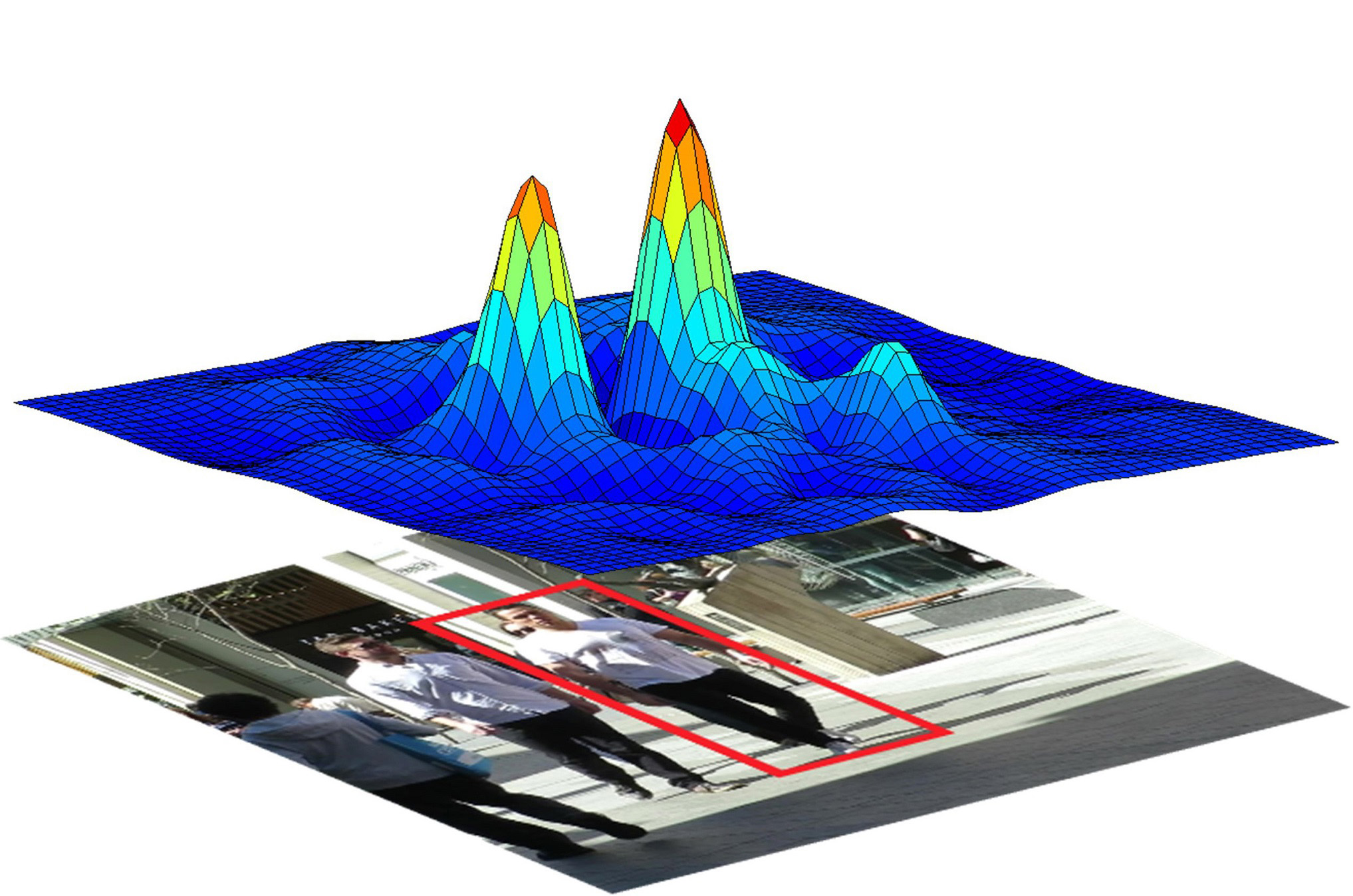}}\\
  \end{minipage}
  \caption{Visualization of the confidence map. The heat map in (a) presents the desired confidence map for the bottom image patch while that in (b) shows the score map predicted by the ECO tracker.}
  \vspace{-4mm}
  \end{figure}
  
  \subsubsection{Cost-Sensitive Tracking Loss.}
  Given an image patch, the ECO tracker utilizes all circular shifted versions of the patch to train the filter. 
  Detection scores of all shifted samples compose the confidence map. Fig.~\ref{fig:class_imbalance_a} shows the desired confidence map for the bottom image patch. 
  The red bounding box in the patch corresponds to the target region.  
  Most locations in the patch are labeled to near zero while only a few locations close to the target center make up positive samples. 
  Fig.~\ref{fig:class_imbalance_b} shows the score map predicted by the ECO tracker. 
  Beside the target location, the center of the object next to the target also gets high confidence score in the middle heat map. 
  Hence, these negative samples centered at intra-class distractors are regarded as hard samples and should be penalized more heavily to prevent the tracker from drifting to the distractor. 
  However, in the ECO formulation \eqref{eq:obj_spatial}, the contributions of all shifted samples in the same search area are weighted equally. 
  Since most negative samples come from the background, the training process may be dominated by substantial background information and consequently degenerate the discriminative power of model on hard samples centered at intra-class distractors.
  
  To alleviate data imbalance, we propose a cost-sensitive loss to put emphasis on hard samples. 
  Specifically, we add a factor $q(t)$ in the data term of \eqref{eq:obj_spatial} as
  \begin{equation}\label{eq:obj_novel}
  E(f)=\sum_{j=1}^M\alpha_j\norm{q(t)(S_f\{\mathbf{x}_j\}(t) - y_j(t))}_{L^2}+\sum_{d=1}^D\norm{w(t)f^d(t)}_{L^2}.
  \end{equation}
  Here, we define the modulating factor $q(t)$ as:
  \begin{equation}\label{eq:q}
  q(t)=\left\lvert\frac{ S_f\{\mathbf{x}_j\}(t) - y_j(t)}{\max_t \left\lvert S_f\{\mathbf{x}_j\}(t) - y_j(t)\right\rvert}\right\rvert^2.
  \end{equation}
  Hence, the modulating factor $q(t)$ re-weights the contributions of circular shifted samples based on their losses.
  
  To make this loss function tractable to solve, we use the filter learned in the last model update step to compute $q(t)$. Thus, $q(t)$ can be precomputed before each training step. 
  Similar to \eqref{eq:obj_spatial}, we transform \eqref{eq:obj_novel} to the objective function in the Fourier domain and perform optimization by solving the following equation:
  \begin{equation}\label{eq:normal_eq_focal}
  \left((\mathbf{QA})^\mathrm{H}\mathbf{\Gamma}(\mathbf{Q} \mathbf{A})+\mathbf{W}^\mathrm{H}\mathbf{W}\right)\,\hat{\mathbf{f}}=(\mathbf{QA})^\mathrm{H}\mathbf{\Gamma}\mathbf{Q}\hat{\mathbf{y}},
  \end{equation}
  where $\mathbf{Q}$ denotes the operation matrix in the Fourier domain, which corresponds to the factor $q(t)$. Like \eqref{eq:normal_eq_eco}, this equation can also be iteratively solved by the Conjugate Gradient
  (CG) method with the same efficiency as the original ECO formulations. Due to the space limit, the concrete derivation and solution  of the proposed cost-sensitive loss are provided in the supplementary material.
  
  \subsection{Data Association with Dual Matching Attention Network}
  When the tracking process becomes unreliable, we suspend the tracker and set the target to a lost state. Then we exploit the data association algorithm to determine whether to keep the target state as lost or transfer it to tracked. It is intuitive to use the tracking score $s$ (i.e., the highest value in the confidence map) of the target to measure the tracking reliability. However, if we only rely on the tracking score, a false alarm detection on the background is prone to be consistently tracked with high confidence. Since a tracked target which does not get any detection for several frames is likely to be a false alarm, we utilize the overlap between bounding boxes given by the tracker and detector to filter out false alarms. Specifically, we set $o(t_l, \mathcal{D}_l)$ to $1$ if the maximum overlap ratio between the tracked target $t_l\in\mathcal{T}_l$ and the detections $\mathcal{D}_l$ in $l$ frames before is higher than 0.5. Otherwise, $o(t_l, \mathcal{D}_l)$ is set to $0$. We consider the mean value of $\{o(t_l, \mathcal{D}_l)\}_1^L$ in the past $L$ tracked frames $o_{\text{mean}}$ as another measurement to decide the tracking state. Thus, the state of the target is defined as:
  \begin{equation}\label{eq:state}
  \text{state}=
  \begin{cases}
  \text{tracked}, & \text{if } s>\tau_s \text{ and } o_{mean}>\tau_o,\\
  \text{lost}, & \text{otherwise}.
  \end{cases}
  \end{equation}
  
  Before computing the appearance similarity for data association, we exploit motion cues to select candidate detections. When the target gets lost, we first keep the scale of the bounding box at the last frame $k-1$ and use a linear motion model to predict its location at the current frame $k$. Denote $\mathbf{c}_{k-1}=[x_{k-1}, y_{k-1}]$ as the center coordinate of the target at frame $k-1$, the velocity $\mathbf{v}_{k-1}$ of the target at frame $k-1$ is computed as:
  \begin{equation}\label{eq:velocity}
  \mathbf{v}_{k-1}=\frac{1}{K}(\mathbf{c}_{k-1} - \mathbf{c}_{k-K}),
  \end{equation}
  where $K$ denotes the frame interval for computing the velocity. Then the target coordinate in the current frame $k$ is predicted as $\tilde{\mathbf{c}}_{k}=\mathbf{c}_{k-1} + \mathbf{v}_{k-1}$.
  
  Given the predicted location of the target, we consider detections surrounding the predicted location which are not covered by any tracked target (i.e., the distance is smaller than a threshold $\tau_d$) as candidate detections. We measure the appearance affinity between these detections and the observations in the target trajectory. Then we select the detection with the highest affinity and set a affinity threshold $\tau_a$ to decide whether to link the lost target to this detection.
  
\begin{figure}[t]
    \centering
    \includegraphics[width=0.9\textwidth]{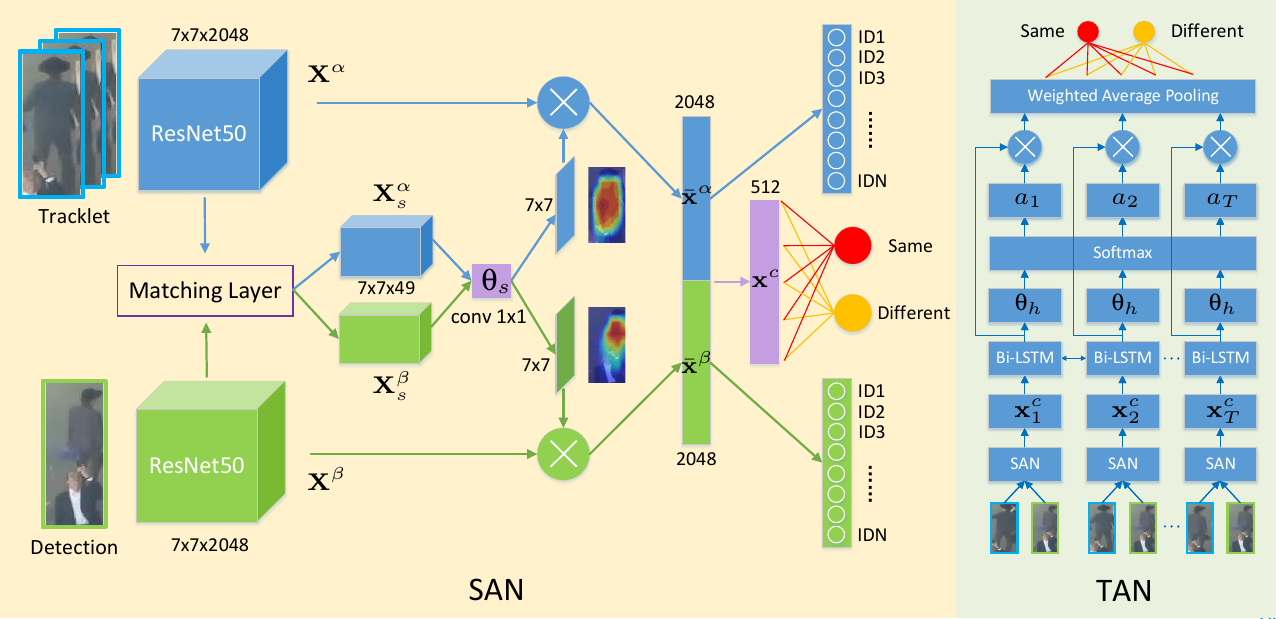}\\
    \caption{Network architecture of the proposed DMAN. It consists of the Spatial Attention Network (SAN) and Temporal Attention Network (TAN). Given a candidate detection and a sequence of the target tracklet as inputs, the SAN repeatedly compares the detection with each sample in the tracklet to extract the combined features $\{\mathbf{x}^c\}_1^T$. Taking these features as inputs, the TAN integrates the information from the overall tracklet to infer whether the detection and the tracklet belong to the same target.}\label{fig:DMAN}
  \vspace{-4mm}
  \end{figure}

  The challenge is that both detections and observations in the tracklet may undergo misalignment and occlusion. To address these problems, we propose Dual Matching Attention Networks (DMAN) with both spatial and temporal attention mechanisms. Fig.~\ref{fig:DMAN} illustrates the architecture of our network.
  \subsubsection{Spatial Attention Network.}
  We propose a spatial attention network using the Siamese architecture to handle noisy detections and occlusions as shown in Fig.~\ref{fig:DMAN}. 
  In this work, we use the truncated ResNet-50 network \cite{ResNet} as the shared base network and apply $L^2$-normalization to output features along the channel dimension. 
  The spatial attention map is applied to the features from the last convolutional layer of the ResNet-50 because representations from the top layer can capture high-level information that is useful for matching semantic regions. 
  We denote the extracted feature map as $\mathbf{X}\in \mathbb{R}^{H\times W\times C}$ and consider $\mathbf{X}$ as a set of $L^2$-normalized $C$-dimension feature vectors:
  \begin{equation}\label{eq:feature_cube}
  \mathbf{X}=\left\{\mathbf{x}_1, \cdots, \mathbf{x}_{N}\right\},\quad \mathbf{x}_i\in\mathbb{R}^C,
  \end{equation}
  where $N=H\times W$. 
  Each feature vector $\mathbf{x}_i$ corresponds to a spatial location on the feature map. 
  Then we denote the feature maps extracted from the image pair as $\mathbf{X}^\alpha=\left\{\mathbf{x}_1^\alpha, \cdots, \mathbf{x}_{N}^\alpha\right\}$ and $\mathbf{X}^\beta=\left\{\mathbf{x}_1^\beta, \cdots, \mathbf{x}_{N}^\beta\right\}$, respectively. 
  The intuition is that we should pay more attention to common local patterns of the two feature maps. 
  However, since the two images are usually not well aligned due to inaccurate bounding boxes and pose change, the corresponding feature located in $\mathbf{X}^\alpha$ may not appears at the same location in $\mathbf{X}^\beta$. 
  Thus, we generate the attention map for each input separately. To infer the attention value for the $i^{th}$ location in the feature map $\mathbf{X}^\alpha$, we need to compare $\mathbf{x}_i^\alpha\in\mathbf{X}^\alpha$ with all the feature slices appearing in the paired feature map $\mathbf{X}^\beta$. 
  
  We exploit a non-parametric matching layer to compute the cosine similarity $S_{ij}=(\mathbf{x}_i^\alpha)^{\top}\mathbf{x}_j^\beta$ between each $\mathbf{x}_i^\alpha$ and $\mathbf{x}_j^\beta$ and output the similarity matrix $\mathbf{S}\in \mathbb{R}^{N\times N}$ as
  \begin{equation}\label{eq:similarity}
  \mathbf{S} =
  \begin{bmatrix}
  (\mathbf{x}_1^\alpha)^{\top}\\
  \vdots\\
  (\mathbf{x}_{N}^\alpha)^{\top}
  \end{bmatrix}
  \cdot\left[\mathbf{x}_1^\beta, \cdots, \mathbf{x}_{N}^\beta\right]
  =
  \begin{bmatrix}
  (\mathbf{s}_1)^{\top}\\
  \vdots\\
  (\mathbf{s}_{N})^{\top}
  \end{bmatrix},
  \end{equation}
  where the vector $\mathbf{s}_i=[S_{i1},\cdots, S_{iN}]^\top\in\mathbb{R}^N$ contains the elements in the $i^{th}$ row of $\mathbf{S}$, which indicate the cosine distances between $\mathbf{x}_i^\alpha\in \mathbf{X}^\alpha$ and all the feature vectors in $\mathbf{X}^\beta$. 
  The similarity matrix $\mathbf{S}\in \mathbb{R}^{N\times N}$ is reshaped into a $H\times W\times N$ feature cube $\mathbf{X}_s^\alpha\in\mathbb{R}^{H\times W\times N}$ to form a similarity representation for the feature map $\mathbf{X}^\alpha$. 
  Then we input $\mathbf{X}_s^\alpha$ to a convolutional layer with $1\times1$ kernel and perform a softmax over the output to generate the attention map $\mathbf{A}^\alpha\in \mathbb{R}^{H\times W}$ for $\mathbf{X}^\alpha$. 
  The attention value $a_i^\alpha$ in $\mathbf{A}^\alpha$ for the $i^{th}$ location in $\mathbf{X}^\alpha$ is defined as:
  \begin{equation}\label{eq:attention}
  a_i^\alpha = \frac{\exp\left(\textnormal{\fontfamily{phv}\straighttheta}_s^{\top}\mathbf{s}_i\right)}{\sum_{i=1}^{N}\exp\left(\textnormal{\fontfamily{phv}\straighttheta}_s^{\top}\mathbf{s}_i\right)},
  \end{equation}
  where $\textnormal{\fontfamily{phv}\straighttheta}_s\in\mathbb{R}^{N}$ denotes the weight of the $1\times 1$ convolutional layer. After applying an average pooling on $\mathbf{X}^\alpha$ weighted by the attention map $\mathbf{A}^\alpha$, we obtain the attention-masked feature $\bar{\mathbf{x}}^\alpha\in \mathbb{R}^C$ as:
  \begin{equation}\label{eq:feature_sp_pooling}
  \bar{\mathbf{x}}^\alpha = \sum_{i=1}^{N}a_i^\alpha\mathbf{x}_i^\alpha.
  \end{equation}
  
  For the feature map $\mathbf{X}^\beta$, we transpose the similarity matrix $\mathbf{S}$ to $\mathbf{S}^{\top}$ so that the $j^{th}$ row of $\mathbf{S}^{\top}$ contains the cosine distances between $\mathbf{x}_j^\beta\in \mathbf{X}^\beta$ and all the feature vectors in $\mathbf{X}^\alpha$. We perform the same operations on $\mathbf{S}^{\top}$ to generate the attention map $\mathbf{A}^\beta\in \mathbb{R}^{H\times W}$ and the masked feature $\bar{\mathbf{x}}^\beta\in \mathbb{R}^C$ for $\mathbf{X}^\beta$. For symmetry, the weights of the $1\times 1$ convolutional layer performed on the similarity representation $\mathbf{X}_s^\alpha, \mathbf{X}_s^\beta$ are shared.
  
  We exploit both the identification loss and verification loss to jointly train the network so that the network needs to simultaneously predict the identity of each image in the input pair and the similarity score between the two images during training. For identification, we apply the cross entropy loss on the masked features $\bar{\mathbf{x}}^\alpha$ and $\bar{\mathbf{x}}^\beta$, respectively. For verification, we concatenate $\bar{\mathbf{x}}^\alpha$ and $\bar{\mathbf{x}}^\beta$ to a single feature and input it to a $512$-dimension fully-connected layer, which outputs the combined feature $\mathbf{x}^c\in\mathbb{R}^{512}$. A binary classifier with cross entropy loss is then performed on $\mathbf{x}^c$ for prediction. 
  
  \subsubsection{Temporal Attention Network.}
  When comparing the candidate detection with a sequence of observations in the tracklet, it is straightforward to apply average pooling on the feature vectors of all the observations in the tracklet for verification. However, as shown in Fig.~\ref{fig:motivation}(b), the tracklet may contain noisy observations. 
  Simply assigning equal weights to all the observations may degrade the model performance. 
  To handle unreliable samples in the tracklet, we exploit the temporal attention mechanism to adaptively allocate different degrees of importance to different samples in the tracklet. Fig.~\ref{fig:DMAN} shows the structure of the proposed temporal attention network.
  
  The temporal attention network takes the set of features $\{\mathbf{x}^c_1,\cdots,\mathbf{x}^c_T\}$  extracted from the spatial attention network as inputs. Here, the feature vector $\mathbf{x}_i^c$ is obtained by comparing the candidate detection with the $i^{th}$ sample in the $T$-length tracklet.
  To determine noisy samples in the tracket, the model should not only rely on the similarity between the detection and each sample in the tracklet (which has been encoded in each $\mathbf{x}^c_i$), but also consider the consistency of all samples. Thus, we utilize a Bi-directional Long-Short Term Memory (Bi-LSTM) network to predict the attention value $a_t$:
  \begin{equation}\label{eq:temporal_att}
  a_t=\frac{\exp\left(\textnormal{\fontfamily{phv}\straighttheta}_h^\top \left[{\mathbf{h}}_t^l;{\mathbf{h}}_t^r\right]\right)}{\sum_{t=1}^{T}\exp\left(\textnormal{\fontfamily{phv}\straighttheta}_h^\top \left[{\mathbf{h}}_t^l;{\mathbf{h}}_t^r\right]\right)},\quad t=1,\cdots,T,
  \end{equation}
  where ${\mathbf{h}}_t^l, {\mathbf{h}}_t^r$ are the bi-directional hidden representations of the Bi-LSTM model and $\textnormal{\fontfamily{phv}\straighttheta}_h$ is the weight of the layer to generate attention values. The attention score $a_t$ is a scalar value which is used to weight the hidden representations $\mathbf{h}_t^l, {\mathbf{h}}_t^r$ of each observation for feature pooling as follows:
  \begin{equation}\label{eq:feature_te_pooling}
  \bar{\mathbf{h}}=\sum_{i=1}^T a_t \left[{\mathbf{h}}_t^l;{\mathbf{h}}_t^r\right].
  \end{equation}
  Taking the pooled feature $\bar{\mathbf{h}}$ as input, the binary classification layer predicts the similarity score between the input detection and paired tracklet.
  
  Finally, we make the assignments between candidate detections and lost targets based on the pairwise similarity scores of detections and tracklets.
  
  \subsubsection{Training Strategy.}
  We utilize the ground-truth detections and identity information provided in the MOT16 training set to generate image pairs and detection-tracklet pairs for network training. However, the training data contains only limited identities and the sequence of each identity consists of consecutive samples with large redundancies. Hence, the proposed network is prone to overfit the training set. To alleviate this problem, we adopt a two-step training strategy. We first train the spatial attention network on randomly generated image pairs. Then we fix the weights of the spatial attention network and use the extracted features as inputs to train the temporal attention network. 
  In addition, we augment the training set by randomly cropping and rescaling the input images.
  To simulate noisy tracklets in practice, we also add noisy samples to the training tracklet sequences by randomly replacing one or two images in the tracklet with images from other identities. 
  Since some targets in the training set contain only a few samples, we randomly sample each identity with the equal probability to alleviate the effect of class imbalance. 
  
  \subsubsection{Trajectory Management.}
  For trajectory initialization, we set a threshold $\tau_{i}$ and discard the target which is lost or not covered by a detection in any of the first $\tau_{i}$ frames. For trajectory termination, we end the target if it keeps lost for over $\tau_{t}$ frames or just exits out of view. In addition, we collect $M$ most recent observations of the target and generate the $T$-length tracklet for data association by uniformly sampling from the collected samples to reduce data redundancy.
  
  \section{Experiments}
  
  \subsubsection{Datasets.}
  We evaluate the proposed online MOT algorithm on the MOT16 \cite{MOT16} and MOT17 benchmark datasets. The MOT16 dataset consists of 14 video sequences (7 for training, 7 for testing). The MOT17 dataset contains the same video sequences as the MOT16 dataset while additionally providing three sets of detections (DPM \cite{DPM}, Faster-RCNN \cite{faster-rcnn}, and SDP \cite{SDP}) for more comprehensive evaluation of the tracking algorithms.
  
  \subsubsection{Evaluation Metrics.}
  We consider the metrics used by the MOT benchmarks \cite{MOT16,MOT15} for evaluation, which includes Multiple Object Tracking Accuracy (MOTA) \cite{clear_mot}, Multiple Object Tracking Precision (MOTP) \cite{clear_mot}, ID F1 score \cite{IDF1} (IDF, the ratio of correct detections over the average number of ground-truth and computed detections), ID Precision \cite{IDF1} (IDP, the fraction of detections that are correctly identified), ID Recall \cite{IDF1} (IDR, the fraction of ground-truth detections that are correctly identified), the ratio of Mostly Tracked targets (MT), the ratio of Mostly Lost targets (ML), the number of False Negatives (FN), the number of False Positives (FP), the number of ID Switches (IDS), the number of fragments (Frag). Note that IDF, IDP, and IDR are recently introduced by Ristani et al. \cite{IDF1} and added to the MOT benchmarks to measure the identity-preserving ability of trackers. We also show the Average Ranking (AR) score suggested by the MOT benchmarks. It is computed by averaging all metric rankings, which can be considered as a reference to compare the overall MOT performance.
  
  \subsubsection{Implementation Details.}
  The proposed method is implemented using MATLAB and Tensorflow \cite{tensorflow}. For single object tracking, we exploit the same features as the ECO-HC \cite{ECO} (i.e., HOG and Color Names). For data association, we use the convolution blocks of the ResNet-50 pre-trained on the ImageNet dataset \cite{ImageNet} as the shared base network. All input images are resized to $224\times224$. The length of the tracklet is set to $T=8$, and the maximum number of collected samples in the trajectory is set to $M=100$. We use the Adam \cite{adam} optimizer to train both the spatial attention network and the temporal attention network. Learning rates of both networks are set to $0.0001$. Let $F$ denote the frame rate of the video, the interval for computing the target velocity is set to $K=0.3F$. The trajectory initialization threshold is set to $\tau_{i}=0.2F$, while the termination threshold is set to $\tau_{t}=2F$. The tracking score threshold is set to $\tau_s=0.2$, and the appearance affinity score threshold is set to $\tau_a=0.6$. All the values of these threshold parameters are set according to the MOTA performance on the MOT16 training set. The source code will be made available to the public.
  
  \vspace{-2mm}
  \subsection{Visualization of the Spatial and Temporal Attention}
  \begin{figure}[t]
  \begin{minipage}[t]{0.5\textwidth}
  \centering
  \subfigure[Spatial attention maps]{
  \label{fig:sp_att}
  \includegraphics[width=0.8\textwidth]{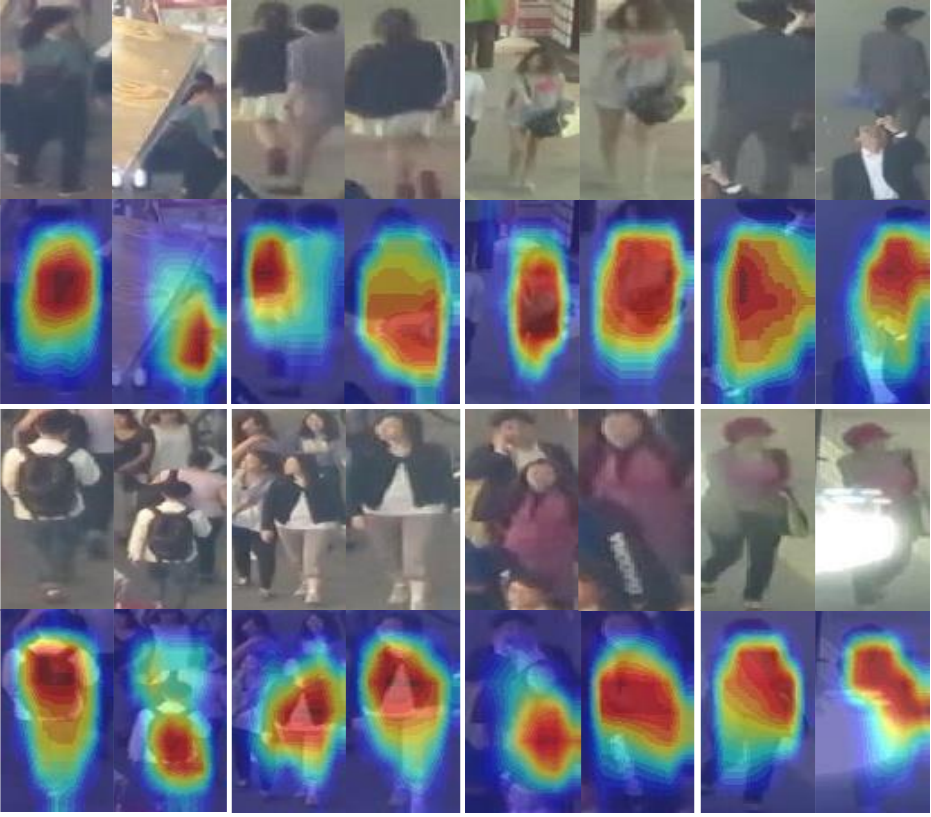}}\\
  \end{minipage}%
  \begin{minipage}[t]{0.5\textwidth}
  \centering
   \subfigure[Temporal attention values]{
    \label{fig:te_att}
    \includegraphics[width=0.8\textwidth]{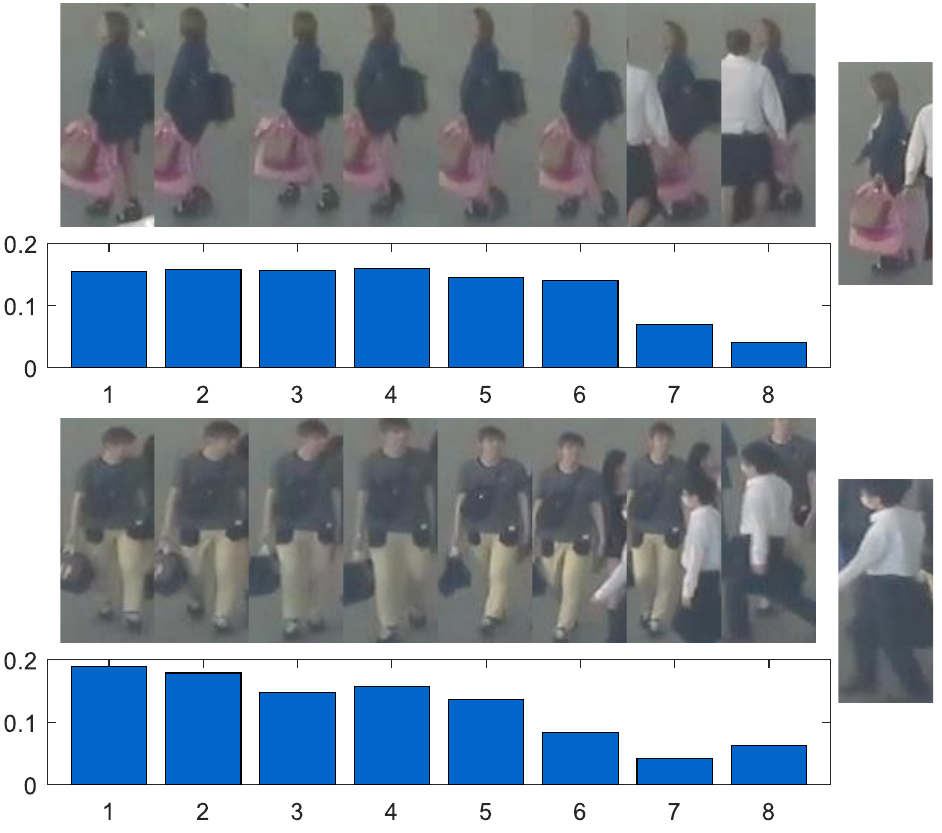}}\\
  \end{minipage}
  \caption{Visualization of spatial and temporal attention.}
  \label{fig:visual_att}
  \vspace{-4mm}
  \end{figure}
  
  Fig.~\ref{fig:visual_att} shows the visualization results of the proposed spatial and temporal attention mechanisms. In  Fig.~\ref{fig:sp_att}, each group consists of four images. The top row of each group shows an image pair from the same target while the bottom row presents corresponding spatial attention maps. Although these image pairs undergo misalignment, scale change, and occlusion, the proposed spatial attention network is still able to locate the matching parts of each pair. Compared with the visibility maps shown in \cite{STAM}, our attention maps focus more explicitly on target regions and suppress both distractors and backgrounds, which enhances the discriminative power of the model on hard positive pairs.
  
  Fig.~\ref{fig:te_att} shows the attention scores predicted by the proposed temporal attention network. The sequence on the left of each row is the tracklet for association while the image on the right of each row corresponds to the candidate detection. The bar chart below the tracklet shows the attention value for each observation. In the top row, the detection and the tracklet belong to the same target. However, the tracklet contains noisy observations caused by occlusion. As shown in the bar chart, the proposed temporal attention network assigns relative low attention scores to occluded observations to suppress their effects on data association. In the bottom row, the detection and the tracklet belong to different targets. Although the last two images in the tracklet contain the same target in the detected patch, the proposed network correctly assigns low attention scores to the last two images by taking the overall sequence into account. These two examples in Fig.~\ref{fig:te_att} demonstrate the effectiveness of the proposed temporal attention mechanism on both hard positive and hard negative samples.
  
  \subsection{Ablation Studies}
  \begin{figure}[t]
    \centering
    \includegraphics[width=0.8\textwidth]{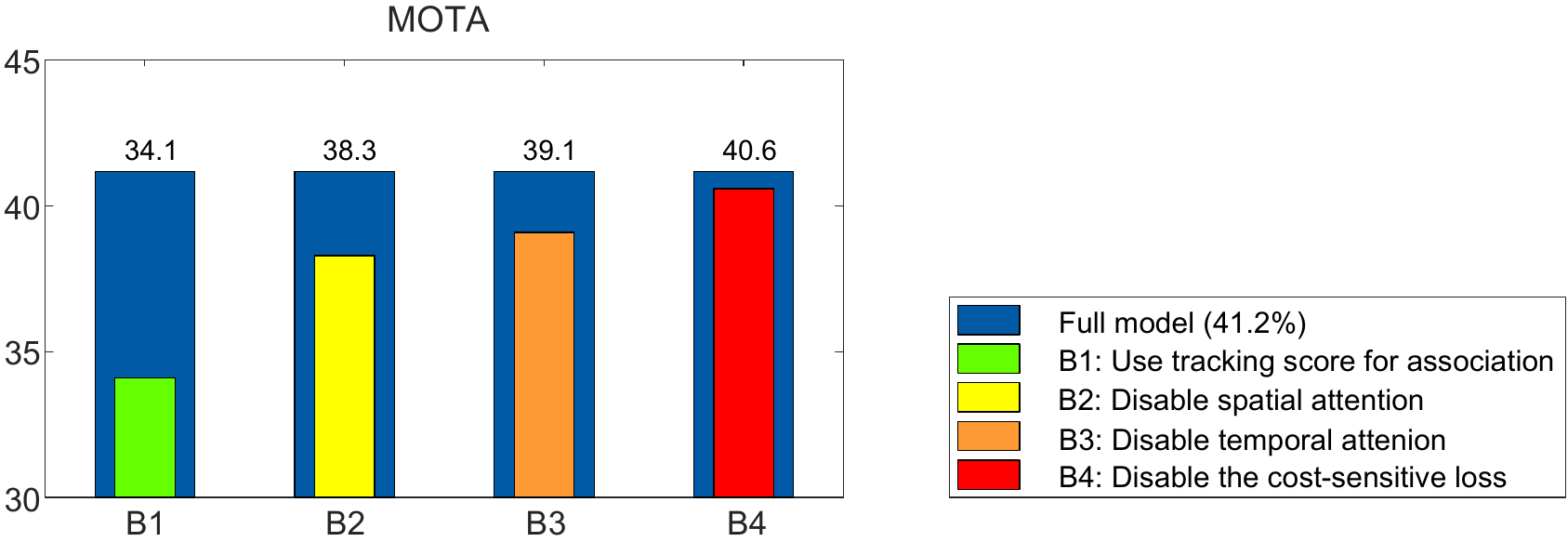}\\
    \caption{Contributions of each component.}\label{fig:baselines}
  \end{figure}
  To demonstrate the contribution of each module in our algorithm, we set up four baseline approaches by disabling each module at one time. Each baseline approach is described as follows:
  
  B1: We disable the proposed DMAN and rely on the cost-sensitive tracker to link the detections. Specifically, we apply the convolution filter of the tracker on the candidate detection and directly use the maximum score in the confidence map as the appearance affinity for data association.
  
  B2: We disable the spatial attention module and use the standard Siamese CNN architecture for identity verification of image pairs.
  
  B3: We replace our temporal attention pooling with average pooling to integrate the hidden representations of the Bi-LSTM in multiple time steps.
  
  B4: We use the baseline tracker without the cost-sensitive tracking loss.
  
  Fig.~\ref{fig:baselines} shows the MOTA score of each baseline approach compared with our full model (41.2\%) on the MOT16 training dataset. As we can see, all proposed modules make contributions to the performance. The MOTA score drops significantly by 7.1\% when we directly use the tracking score for data association, which shows the advantage of the proposed DMAN. The degradation in B2 and B3 demonstrates the effectiveness of the proposed attention mechanisms. Finally, the cost-sensitive tracking loss shows a slight improvement in term of MOTA.
  
  \subsection{Performance on the MOT Benchmark Datasets}
  We evaluate our approach on the test sets of both the MOT16 and MOT17 benchmark against the state-of-the-art methods. Table~\ref{tab:mot16} and Table~\ref{tab:mot17} present the quantitative performance on the MOT16 and MOT17 datasets, respectively.
  
  As shown in Table~\ref{tab:mot16}, our method achieves a comparable MOTA score and performs favorably against the state-of-the-art methods in terms of IDF, IDP, IDR, MT, and FN on the MOT16 dataset. We improve 4.8\% in IDF, 3.9\% in IDP, 4\% in IDR, and 2.8\% in MT compared with the second best published online MOT tracker and achieves the best performance in IDF and IDP among both online and offline methods, which demonstrates the merits of our approach in maintaining identity. Similarly, Table~\ref{tab:mot17} shows that the proposed method performs favorably against the other online trackers in MOTA and achieves the best performance in terms of identity-preserving metrics (IDF, IDP, IDR, IDS) among all methods on the MOT17 dataset. In addition, we achieve the best AR score among all the online trackers on both the MOT16 and MOT17 datasets.
  \begin{table*}[!t]
    \caption{Tracking performance on the MOT16 dataset. }
    \label{tab:mot16}
    \centering
    \tiny
    \begin{tabular}{@{}L{0.8cm}L{1.8cm}C{0.8cm}C{0.8cm}C{0.6cm}C{0.6cm}C{0.6cm}R{0.6cm}R{0.7cm}R{0.75cm}R{0.9cm}R{0.6cm}R{0.6cm}R{0.5cm}@{}}
      \toprule
      \multicolumn{1}{c}{Mode} & \multicolumn{1}{c}{Method} & \multicolumn{1}{c}{MOTA\,$\uparrow$} & \multicolumn{1}{c}{MOTP\,$\uparrow$} & \multicolumn{1}{c}{IDF\,$\uparrow$} & \multicolumn{1}{c}{IDP\,$\uparrow$} & \multicolumn{1}{c}{IDR\,$\uparrow$} & \multicolumn{1}{c}{MT\,$\uparrow$} & \multicolumn{1}{c}{ML\,$\downarrow$} & \multicolumn{1}{c}{FP\,$\downarrow$} & \multicolumn{1}{c}{FN\,$\downarrow$} & \multicolumn{1}{c}{IDS\,$\downarrow$} & \multicolumn{1}{c}{Frag\,$\downarrow$} &  \multicolumn{1}{c}{AR\,$\downarrow$}\\
      \midrule
    \multirow{7}{*}{Online}   & OVBT \cite{OVBT}         & 38.4 & 75.4 & 37.8 & 55.4 & 28.7 &  7.5\% & 47.3\% & 11,517 &  99,463 & 1,321 & 2,140 & 49.8\\
                              & EAMTT \cite{EAMTT}        & 38.8 & 75.1 & 42.4 & 65.2 & 31.5 &  7.9\% & 49.1\% &  8,114 & 102,452 &   965 & 1,657 & 37.4\\
                              & oICF \cite{oICF}          & 43.2 & 74.3 & 49.3 & 73.3 & 37.2 & 11.3\% & 48.5\% &  6,651 &  96,515 &   \textbf{\red381} & \textbf{\red1,404} & 33.3\\
                              & CDA\_DDAL \cite{CDA_DDAL}     & 43.9 & 74.7 & 45.1 & 66.5 & 34.1 & 10.7\% & 44.4\% &  6,450 &  95,175 &   676 & 1,795 & 31.8\\
                              & STAM \cite{STAM}          & 46.0 & 74.9 & 50.0 & 71.5 & 38.5 & 14.6\% & 43.6\% &  6,895 &  91,117 &   473 & 1,422 & 29.6\\
                              & AMIR \cite{AMIR}          & \textbf{\red47.2} & \textbf{\red75.8} & 46.3 & 68.9 & 34.8 & 14.0\% & \textbf{\red41.6\%} &  \textbf{\red2,681} &  92,856 &   774 & 1,675 & 21.8\\
                              & \textbf{Ours} & 46.1 & 73.8 & \textbf{\red54.8} & \textbf{\red77.2} & \textbf{\red42.5} & \textbf{\red17.4\%} & 42.7\% &  7,909 &  \textbf{\red89,874} &   532 & 1,616 & \textbf{\red19.3}\\
    \midrule
    \multirow{10}{*}{Offline}
                              & QuadMOT \cite{QuadMOT}       & 44.1 & 76.4 & 38.3 & 56.3 & 29.0 & 14.6\% & 44.9\% &  6,388 &  94,775 &   745 &  1,096 & 31.9\\
                              & EDMT \cite{EDMT}          & 45.3 & 75.9 & 47.9 & 65.3 & 37.8 & 17.0\% & 39.9\% & 11,122 &  87,890 &   639 &   946 & 20.3\\
                              & MHT\_DAM \cite{MHT_DAM}      & 45.8 & 76.3 & 46.1 & 66.3 & 35.3 & 16.2\% & 43.2\% &  6,412 &  91,758 &   590 &   781 & 23.7\\
                              & JMC \cite{JMC}           & 46.3 & 75.7 & 46.3 & 66.3 & 35.6 & 15.5\% & \textbf{\red39.7\%} &  6,373 &  90,914 &   657 & 1,114 & 21.1\\
                              & NOMT \cite{NOMT}          & 46.4 & 76.6 & \textbf{\red53.3} & \textbf{\red73.2} & \textbf{\red41.9} & 18.3\% & 41.4\% &  9,753 &  87,565 &   \textbf{\red359} &   \textbf{\red504} & 16.3\\
                              & MCjoint \cite{MCjoint}       & 47.1 & 76.3 & 52.3 & 73.9 & 40.4 & \textbf{\red20.4\%} & 46.9\% &  6,703 &  89,368 &   370 &   598 & 18.6\\
                              & NLLMPa \cite{NLLMPa}        & 47.6 & 78.5 & 47.3 & 67.2 & 36.5 & 17.0\% & 40.4\% &  \textbf{\red5,844} &  89,093 &   629 &   768 & 16.8\\
                              & LMP \cite{LMP}           & \textbf{\red48.8} & \textbf{\red79.0} & 51.3 & 71.1 & 40.1 & 18.2\% & 40.1\% &  6,654 &  \textbf{\red86,245} &   481 &   595 & \textbf{\red14.8}\\
      \bottomrule
    \end{tabular}
  \end{table*}
  \begin{table*}[t]
    \caption{Tracking performance on the MOT17 dataset. } 
    \label{tab:mot17}
    \centering
    \tiny
    \begin{tabular}{@{}L{0.8cm}L{2.0cm}C{0.8cm}C{0.8cm}C{0.6cm}C{0.6cm}C{0.6cm}R{0.6cm}R{0.7cm}R{0.75cm}R{0.9cm}R{0.6cm}R{0.6cm}R{0.5cm}@{}}
      \toprule
      \multicolumn{1}{c}{Mode} & \multicolumn{1}{c}{Method} & \multicolumn{1}{c}{MOTA\,$\uparrow$} & \multicolumn{1}{c}{MOTP\,$\uparrow$} & \multicolumn{1}{c}{IDF\,$\uparrow$} & \multicolumn{1}{c}{IDP\,$\uparrow$} & \multicolumn{1}{c}{IDR\,$\uparrow$} & \multicolumn{1}{c}{MT\,$\uparrow$} & \multicolumn{1}{c}{ML\,$\downarrow$} & \multicolumn{1}{c}{FP\,$\downarrow$} & \multicolumn{1}{c}{FN\,$\downarrow$} & \multicolumn{1}{c}{IDS\,$\downarrow$} & \multicolumn{1}{c}{Frag\,$\downarrow$} &  \multicolumn{1}{c}{AR\,$\downarrow$}\\
      \midrule
    \multirow{5}{*}{Online}
                            & GM\_PHD \cite{GM_PHD}       & 36.4 & 76.2 & 33.9 & 54.2 & 24.7 &  4.1\% & 57.3\% &  23,723 & 330,767 & 4,607 & 11,317 & 23.0\\
                            & GMPHD\_KCF \cite{GMPHD_KCF}  & 39.6 & 74.5 & 36.6 & 49.6 & 29.1 &  8.8\% & 43.3\% & 50,903 & 284,228 & 5,811 &  7,414 & 23.5\\
                            & E2EM  & 47.5 & \textbf{\red76.5} & 48.8 & 68.4 & 37.9 & 16.5\% &  \textbf{\red37.5\%}  & \textbf{\red20,655} & 272,187 & 3,632 &  12,712 & 13.1\\
                            & \textbf{Ours} & \textbf{\red48.2} &  75.9  & \textbf{\red55.7} & \textbf{\red75.9}     & \textbf{\red44.0}     & \textbf{\red19.3\%} & 38.3\% &  26,218 & \textbf{\red263,608} & \textbf{\red2,194} &  \textbf{\red5,378} & \textbf{\red11.4}\\
    \midrule
    \multirow{3}{*}{Offline}  & IOU \cite{IOU}      & 45.5 & 76.9 & 39.4 & 56.4 & 30.3 & 15.7\% & 40.5\% &  \textbf{\red19,993} & 281,643 & 5,988 &  7,404 & 16.4\\
                               & EDMT \cite{EDMT}     & 50.0 & 77.3 & \textbf{\red51.3} & \textbf{\red67.0} & \textbf{\red41.5} & \textbf{\red21.6\%} & \textbf{\red36.3\%} &  32,279 & \textbf{\red247,297} & \textbf{\red2,264} &  3,260 &  \textbf{\red9.9}\\
                               & MHT\_DAM\cite{MHT_DAM} & \textbf{\red50.7} & \textbf{\red77.5} & 47.2 & 63.4 & 37.6 & 20.8\% & 36.9\% &  22,875 & 252,889 & 2,314 &  \textbf{\red2,865} & 10.8\\
      \bottomrule
    \end{tabular}
  \end{table*}
  \vspace{-3mm}
  \section{Conclusions}
  In this work, we integrate the merits of single object tracking and data association methods in a unified online MOT framework. 
  For single object tracking, we introduce a novel cost-sensitive loss to mitigate the effects of data imbalance. 
  For data association, we exploit both the spatial and temporal attention mechanisms to handle noisy detections and occlusions. 
  Experimental results on public MOT benchmark datasets demonstrate the effectiveness of the proposed approach.
  
  \subsubsection{Acknowledgments.}
  This work is supported in part by National Natural Science Foundation of China (NSFC, Grant No. 61771303, 61671289, and 61521062), Science and Technology Commission of Shanghai Municipality (STCSM, Grant No. 17DZ1205602, 18DZ1200102, and 18DZ2270700), SJTU-YITU/Thinkforce Joint Lab of Visual Computing and Application, and Visbody. J. Zhu and N. Liu are supported by a scholarship from China Scholarship Council. M. Kim is supported by the Panasonic Silicon Valley Laboratory. M.-H. Yang acknowlegdes the support from NSF (Grant No. 1149783) and gifts from Adobe and NVIDIA. 
  \clearpage
  
  \bibliographystyle{splncs04}
  \bibliography{0971}

\begin{thebibliography}{10}
\providecommand{\url}[1]{\texttt{#1}}
\providecommand{\urlprefix}{URL }
\providecommand{\doi}[1]{https://doi.org/#1}

\bibitem{tensorflow}
Abadi, M., Agarwal, A., Barham, P., Brevdo, E., Chen, Z., Citro, C., Corrado,
  G.S., Davis, A., Dean, J., Devin, M., et~al.: Tensorflow: Large-scale machine
  learning on heterogeneous distributed systems. arXiv preprint
  arXiv:1603.04467  (2016)

\bibitem{CDA_DDAL}
Bae, S.H., Yoon, K.J.: Confidence-based data association and discriminative
  deep appearance learning for robust online multi-object tracking. TPAMI
  (2017)

\bibitem{OVBT}
Ban, Y., Ba, S., Alameda-Pineda, X., Horaud, R.: Tracking multiple persons
  based on a variational bayesian model. In: ECCV Workshop (2016)

\bibitem{clear_mot}
Bernardin, K., Stiefelhagen, R.: Evaluating multiple object tracking
  performance: the {CLEAR MOT} metrics. JIVP  (2008)

\bibitem{IOU}
Bochinski, E., Eiselein, V., Sikora, T.: High-speed tracking-by-detection
  without using image information. In: AVSS Workshop (2017)

\bibitem{bulo2017loss}
Bulo, S.R., Neuhold, G., Kontschieder, P.: Loss max-pooling for semantic image
  segmentation. In: CVPR (2017)

\bibitem{EDMT}
Chen, J., Sheng, H., Zhang, Y., Xiong, Z.: Enhancing detection model for
  multiple hypothesis tracking. In: CVPR Workshop (2017)

\bibitem{chen2015mind}
Chen, X., Lawrence~Zitnick, C.: Mind's eye: A recurrent visual representation
  for image caption generation. In: CVPR (2015)

\bibitem{NOMT}
Choi, W.: Near-online multi-target tracking with aggregated local flow
  descriptor. In: ICCV (2015)

\bibitem{STAM}
Chu, Q., Ouyang, W., Li, H., Wang, X., Liu, B., Yu, N.: Online multi-object
  tracking using cnn-based single object tracker with spatial-temporal
  attention mechanism. In: ICCV (2017)

\bibitem{HOG}
Dalal, N., Triggs, B.: Histograms of oriented gradients for human detection.
  In: CVPR (2005)

\bibitem{ECO}
Danelljan, M., Bhat, G., Khan, F.S., Felsberg, M.: {ECO}: Efficient convolution
  operators for tracking. In: CVPR (2017)

\bibitem{CCOT}
Danelljan, M., Robinson, A., Khan, F.S., Felsberg, M.: Beyond correlation
  filters: Learning continuous convolution operators for visual tracking. In:
  ECCV (2016)

\bibitem{dehghan2015target}
Dehghan, A., Tian, Y., Torr, P.H., Shah, M.: Target identity-aware network flow
  for online multiple target tracking. In: CVPR (2015)

\bibitem{ImageNet}
Deng, J., Dong, W., Socher, R., Li, L.J., Li, K., Fei-Fei, L.: Imagenet: A
  large-scale hierarchical image database. In: CVPR (2009)

\bibitem{GM_PHD}
Eiselein, V., Arp, D., P{\"a}tzold, M., Sikora, T.: Real-time multi-human
  tracking using a probability hypothesis density filter and multiple
  detectors. In: AVSS (2012)

\bibitem{fang2015captions}
Fang, H., Gupta, S., Iandola, F., Srivastava, R.K., Deng, L., Doll{\'a}r, P.,
  Gao, J., He, X., Mitchell, M., Platt, J.C., et~al.: From captions to visual
  concepts and back. In: CVPR (2015)

\bibitem{felzenszwalb2010cascade}
Felzenszwalb, P.F., Girshick, R.B., McAllester, D.: Cascade object detection
  with deformable part models. In: CVPR (2010)

\bibitem{DPM}
Felzenszwalb, P.F., Girshick, R.B., McAllester, D., Ramanan, D.: Object
  detection with discriminatively trained part-based models. TPAMI
  \textbf{32}(9),  1627--1645 (2010)

\bibitem{ResNet}
He, K., Zhang, X., Ren, S., Sun, J.: Deep residual learning for image
  recognition. In: CVPR (2016)

\bibitem{MCjoint}
Keuper, M., Tang, S., Zhongjie, Y., Andres, B., Brox, T., Schiele, B.: A
  multi-cut formulation for joint segmentation and tracking of multiple
  objects. arXiv preprint arXiv:1607.06317  (2016)

\bibitem{oICF}
Kieritz, H., Becker, S., H{\"u}bner, W., Arens, M.: Online multi-person
  tracking using integral channel features. In: AVSS (2016)

\bibitem{MHT_DAM}
Kim, C., Li, F., Ciptadi, A., Rehg, J.M.: Multiple hypothesis tracking
  revisited. In: ICCV (2015)

\bibitem{adam}
Kingma, D., Ba, J.: Adam: A method for stochastic optimization. arXiv preprint
  arXiv:1412.6980  (2014)

\bibitem{VOT}
Kristan, M., Matas, J., Leonardis, A., Felsberg, M., Cehovin, L.,
  Fern{\'a}ndez, G., Vojir, T., Hager, G., Nebehay, G., Pflugfelder, R.: The
  visual object tracking {VOT2015} challenge results. In: ECCV Workshop (2015)

\bibitem{GMPHD_KCF}
Kutschbach, T., Bochinski, E., Eiselein, V., Sikora, T.: Sequential sensor
  fusion combining probability hypothesis density and kernelized correlation
  filters for multi-object tracking in video data. In: AVSS (2017)

\bibitem{leal2016learning}
Leal-Taix{\'e}, L., Canton-Ferrer, C., Schindler, K.: Learning by tracking:
  Siamese cnn for robust target association. In: CVPR Workshop (2016)

\bibitem{MOT15}
Leal-Taix{\'e}, L., Milan, A., Reid, I., Roth, S., Schindler, K.: {MOTchallenge
  2015}: Towards a benchmark for multi-target tracking. arXiv preprint
  arXiv:1504.01942  (2015)

\bibitem{NLLMPa}
Levinkov, E., Uhrig, J., Tang, S., Omran, M., Insafutdinov, E., Kirillov, A.,
  Rother, C., Brox, T., Schiele, B., Andres, B.: Joint graph decomposition \&
  node labeling: Problem, algorithms, applications. In: CVPR (2017)

\bibitem{CUHK02}
Li, W., Wang, X.: Locally aligned feature transforms across views. In: CVPR
  (2013)

\bibitem{CUHK01}
Li, W., Zhao, R., Wang, X.: Human reidentification with transferred metric
  learning. In: ACCV (2012)

\bibitem{CUHK03}
Li, W., Zhao, R., Xiao, T., Wang, X.: Deepreid: Deep filter pairing neural
  network for person re-identification. In: CVPR (2014)

\bibitem{TC}
Liang, P., Blasch, E., Ling, H.: Encoding color information for visual
  tracking: Algorithms and benchmark. TIP  \textbf{24}(12),  5630--5644 (2015)

\bibitem{focal_loss}
Lin, T.Y., Goyal, P., Girshick, R., He, K., Doll{\'a}r, P.: Focal loss for
  dense object detection. In: ICCV (2017)

\bibitem{MOT16}
Milan, A., Leal-Taix{\'e}, L., Reid, I., Roth, S., Schindler, K.: {MOT16}: A
  benchmark for multi-object tracking. arXiv preprint arXiv:1603.00831  (2016)

\bibitem{milan2017online}
Milan, A., Rezatofighi, S.H., Dick, A.R., Reid, I.D., Schindler, K.: Online
  multi-target tracking using recurrent neural networks. In: AAAI (2017)

\bibitem{milan2014continuous}
Milan, A., Roth, S., Schindler, K.: Continuous energy minimization for
  multitarget tracking. TPAMI  \textbf{36}(1),  58--72 (2014)

\bibitem{UAV}
Mueller, M., Smith, N., Ghanem, B.: A benchmark and simulator for uav tracking.
  In: ECCV (2016)

\bibitem{pirsiavash2011globally}
Pirsiavash, H., Ramanan, D., Fowlkes, C.C.: Globally-optimal greedy algorithms
  for tracking a variable number of objects. In: CVPR (2011)

\bibitem{faster-rcnn}
Ren, S., He, K., Girshick, R., Sun, J.: Faster r-cnn: Towards real-time object
  detection with region proposal networks. In: NIPS (2015)

\bibitem{IDF1}
Ristani, E., Solera, F., Zou, R., Cucchiara, R., Tomasi, C.: Performance
  measures and a data set for multi-target, multi-camera tracking. In: ECCV
  Workshop (2016)

\bibitem{AMIR}
Sadeghian, A., Alahi, A., Savarese, S.: Tracking the untrackable: Learning to
  track multiple cues with long-term dependencies. In: ICCV (2017)

\bibitem{EAMTT}
Sanchez-Matilla, R., Poiesi, F., Cavallaro, A.: Multi-target tracking with
  strong and weak detections. In: ECCV Workshop (2016)

\bibitem{shrivastava2016training}
Shrivastava, A., Gupta, A., Girshick, R.: Training region-based object
  detectors with online hard example mining. In: CVPR (2016)

\bibitem{QuadMOT}
Son, J., Baek, M., Cho, M., Han, B.: Multi-object tracking with quadruplet
  convolutional neural networks. In: CVPR (2017)

\bibitem{tang2015subgraph}
Tang, S., Andres, B., Andriluka, M., Schiele, B.: Subgraph decomposition for
  multi-target tracking. In: CVPR (2015)

\bibitem{JMC}
Tang, S., Andres, B., Andriluka, M., Schiele, B.: Multi-person tracking by
  multicut and deep matching. In: ECCV Workshop (2016)

\bibitem{LMP}
Tang, S., Andriluka, M., Andres, B., Schiele, B.: Multiple people tracking by
  lifted multicut and person re-identification. In: CVPR (2017)

\bibitem{CN}
Van De~Weijer, J., Schmid, C., Verbeek, J., Larlus, D.: Learning color names
  for real-world applications. TIP  (2009)

\bibitem{wang2017residual}
Wang, F., Jiang, M., Qian, C., Yang, S., Li, C., Zhang, H., Wang, X., Tang, X.:
  Residual attention network for image classification. In: CVPR (2017)

\bibitem{wang2016tracking}
Wang, X., T{\"u}retken, E., Fleuret, F., Fua, P.: Tracking interacting objects
  using intertwined flows. TPAMI  \textbf{38}(11),  2312--2326 (2016)

\bibitem{OTB}
Wu, Y., Lim, J., Yang, M.H.: Object tracking benchmark. TPAMI  \textbf{37}(9),
  1834--1848 (2015)

\bibitem{MDP}
Xiang, Y., Alahi, A., Savarese, S.: Learning to track: Online multi-object
  tracking by decision making. In: ICCV (2015)

\bibitem{xu2016ask}
Xu, H., Saenko, K.: Ask, attend and answer: Exploring question-guided spatial
  attention for visual question answering. In: ECCV (2016)

\bibitem{xu2015show}
Xu, K., Ba, J., Kiros, R., Cho, K., Courville, A., Salakhudinov, R., Zemel, R.,
  Bengio, Y.: Show, attend and tell: Neural image caption generation with
  visual attention. In: ICML (2015)

\bibitem{SDP}
Yang, F., Choi, W., Lin, Y.: Exploit all the layers: Fast and accurate cnn
  object detector with scale dependent pooling and cascaded rejection
  classifiers. In: CVPR (2016)

\bibitem{yang2016stacked}
Yang, Z., He, X., Gao, J., Deng, L., Smola, A.: Stacked attention networks for
  image question answering. In: CVPR (2016)

\bibitem{zhang2008global}
Zhang, L., Li, Y., Nevatia, R.: Global data association for multi-object
  tracking using network flows. In: CVPR (2008)

\end{thebibliography}
  \end{document}